\documentclass[10pt,twocolumn,letterpaper]{article}

\usepackage{cvpr}
\usepackage{times}
\usepackage{epsfig}
\usepackage{graphicx}
\usepackage{amsmath}
\usepackage{amssymb}

\usepackage[breaklinks=true,bookmarks=false]{hyperref}

\cvprfinalcopy 

\def\cvprPaperID{****} 

\ifcvprfinal\fi

\begin{document}

\title{Incorporating Copying Mechanism in Image Captioning \\for Learning Novel Objects}

\author{Ting Yao, Yingwei Pan, Yehao Li, and Tao Mei\\
Microsoft Research, Beijing, China\\
University of Science and Technology of China, Hefei, China\\
Sun Yat-Sen University, Guangzhou, China\\
{\tt\small \{tiyao, tmei\}@microsoft.com, \{panyw.ustc, yehaoli.sysu\}@gmail.com}}

\maketitle
\thispagestyle{empty}

\begin{abstract}
Image captioning often requires a large set of training image-sentence pairs. In practice, however, acquiring sufficient training pairs is always expensive, making the recent captioning models limited in their ability to describe objects outside of training corpora (i.e., novel objects). In this paper, we present Long Short-Term Memory with Copying Mechanism (LSTM-C) --- a new architecture that incorporates copying into the Convolutional Neural Networks (CNN) plus Recurrent Neural Networks (RNN) image captioning framework, for describing novel objects in captions. Specifically, freely available object recognition datasets are leveraged to develop classifiers for novel objects. Our LSTM-C then nicely integrates the standard word-by-word sentence generation by a decoder RNN with copying mechanism which may instead select words from novel objects at proper places in the output sentence. Extensive experiments are conducted on both MSCOCO image captioning and ImageNet datasets, demonstrating the ability of our proposed LSTM-C architecture to describe novel objects. Furthermore, superior results are reported when compared to state-of-the-art deep models.
\end{abstract}

\section{Introduction}
\begin{figure}[!tb]
   \centering {\includegraphics[width=0.49\textwidth]{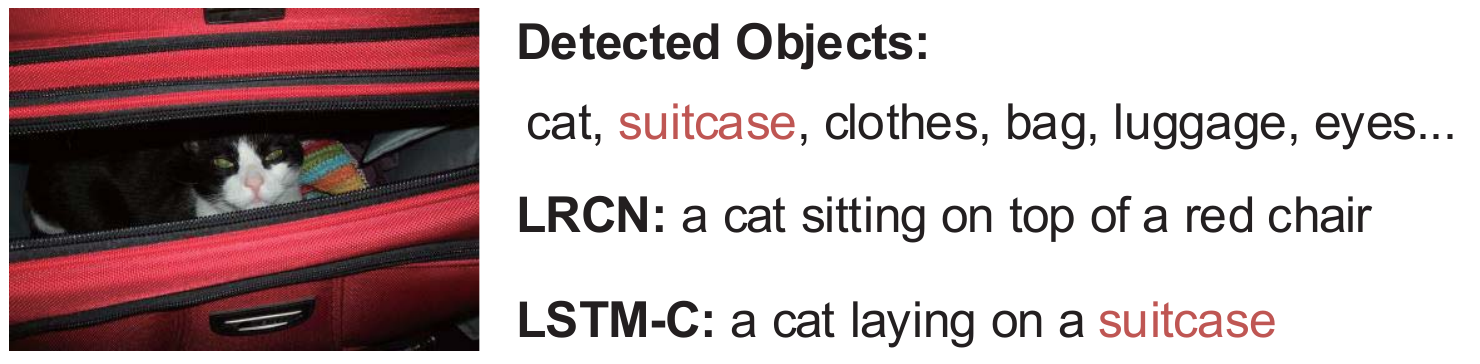}}
   \caption{\small An example of object recognition and image captioning. The input is an image, while the output is the detected objects and a natural sentence, respectively. (upper row: the detected objects in the image; middle row: the sentence generated by LRCN \cite{Donahue14} image captioning approach; bottom row: the sentence generated by our LSTM-C model.)}
   \label{fig:fig1}
   \vspace{-0.15in}
\end{figure}

Automatically describing the content of an image with a complete and natural sentence, a problem known as image captioning, has great potential impact for instance on robotic vision or helping visually impaired people. Intensive research interests from both computer vision and natural language processing communities have been paid for this emerging topic. Most of recent attempts on this problem \cite{Donahue14,Vinyals14,Xu:ICML15,You:CVPR16} are Convolutional Neural Networks (CNN) plus Recurrent Neural Networks (RNN) based sequence learning methods, which are mainly inspired from the advances by using RNN in machine translation \cite{Sutskever:NIPS14}. The basic idea is an encoder-decoder mechanism for translation. Specifically, a CNN is employed to encode image content and then a decoder RNN is exploited to generate a natural sentence. While encouraging performances are reported, the sequence learning methods learn directly from image and sentence pairs, which fail in their ability to describe the objects out of the training data, i.e., novel objects. Take the image in Figure \ref{fig:fig1} as an example, the output sentence generated by a popular image captioning method Long-term Recurrent Convolutional Networks (LRCN) \cite{Donahue14} is unable to describe ``suitcase" as this object is non-existent in the training corpora. More importantly, manually labeling a large-scale image captioning dataset is an intellectually expensive and time-consuming process.

We demonstrate in this paper that the above limitations could be mitigated by incorporating the knowledge from external visual recognition datasets, which are freely available for developing object detectors. Specifically, we present a novel Long Short-Term Memory with Copying Mechanism (LSTM-C) framework to generate words by integrating ``copying mechanism." Copying mechanism is originated from human language communication and refers to the mechanism that locates a certain segment of the input sequence and directly puts the segment in the output sequence \cite{gu2016incorporating}. The spirit behind is the rote memorization in language processing of human being, which needs to refer to sub-sequences of the input. We extend the copying idea here to select novel objects learnt from external sources and put them at proper places in the generated sentence. The overview of LSTM-C framework is illustrated in Figure \ref{fig:figLC}. Given an image, a CNN is utilized to extract visual features, which will be fed into LSTM at the initial time step for sentence generation. Meanwhile, the objects of the input image are also predicted by object detectors pre-trained on recognition dataset. A copying layer is devised at the top of the whole architecture to accommodate the generative model of LSTM and copying mechanism from the detected objects. By integrating copying mechanism into image captioning, the word ``suitcase" is copied from detected objects and output in the sentence generated by our LSTM-C as shown in Figure \ref{fig:fig1}. The whole architecture is trained end-to-end.

The main contribution of this work is the proposal of LSTM-C framework by incorporating the knowledge from external sources to address the issue of predicting novel objects in image captioning task. This issue also leads to an elegant view of how to accommodate both generative model and copying mechanism from detected objects for sentence generation, which is a problem not yet fully understood.

\section{Related Work}\label{sec:RW}
We briefly group the related works into two categories: image captioning and novel object captioning. The first category reviews the research in sentence generation for images, while the second investigates a variety of recent models which attempt to describe novel objects in context.

\subsection{Image Captioning}
The research on image captioning has proceeded along three different dimensions: template-based methods \cite{Kulkarni:PAMI13,Mitchell:EACL12,Yang:EMNLP11}, search-based approaches \cite{devlin2015language,Farhadi:ECCV10,Ordonez:NIPS11}, and language-based models \cite{Donahue14,Kiros:ICML14,Vinyals14,Wu:CVPR16,Xu:ICML15,Yao:arxiv16,You:CVPR16}.

Template-based methods predefine the template for sentence generation and split sentence into several parts (e.g., subject, verb, and object). With such sentence fragments, many works align each part with visual content (e.g., CRF in \cite{Kulkarni:PAMI13} and HMM in \cite{Yang:EMNLP11}) and then generate the sentence for the image. Obviously, most of them highly depend on the templates of sentence and always generate sentence with syntactical structure. Search-based approaches \cite{devlin2015language,Farhadi:ECCV10,Ordonez:NIPS11} ``generate" sentence for an image by selecting the most semantically similar sentences from sentence pool. This direction indeed can achieve human-level descriptions as all the output sentences are from existing human-generated ones. The need to collect human-generated sentences, however, makes the sentence pool hard to be scaled~up.

Different from template-based and search-based models, language-based models aim to learn the probability distribution in the common space of visual content and textual sentence to generate novel sentences with more flexible syntactical structures. In this direction, recent works explore such probability distribution mainly using neural networks and have achieved promising results for image captioning task. Kiros \emph{et al.} \cite{Kiros:ICML14} employ the neural networks to generate sentence for an image by proposing a multimodal log-bilinear neural language model. In \cite{Vinyals14}, Vinyals \emph{et al.} propose an end-to-end neural networks architecture by utilizing LSTM to generate sentence for an image, which is further incorporated with attention mechanism in \cite{Xu:ICML15} to automatically focus on salient objects when generating corresponding words. More recently, in \cite{Wu:CVPR16}, high-level concepts/attributes are shown to obtain clear improvements on image captioning task when injected into existing state-of-the-art RNN-based model. Such high-level attributes are further utilized as semantic attention in \cite{You:CVPR16} and complementary representations to visual features in \cite{Pan:arxiv16,Yao:arxiv16} to enhance image/video captioning.

\subsection{Novel Object Captioning}
The novel object captioning is a new problem that has received increasing attention most recently, which leverages additional image-sentence paired data \cite{Mao:ICCV15} or unpaired image/text data \cite{Hendricks:CVPR16,venugopalan2016captioning} to describe novel objects in existing RNN-based image captioning frameworks. \cite{Mao:ICCV15} is one of the early works that enlarges the original limited word dictionary to describe novel objects by using only a few paired image-sentence data. In particular, a transposed weight sharing scheme is proposed to avoid extensive retraining. In contrast, with the largely available unpaired image/text data (e.g., ImageNet and Wikipedia), Hendricks \emph{et al.} \cite{Hendricks:CVPR16} explicitly transfer the knowledge of semantically related objects to compose the descriptions about novel objects in the proposed Deep Compositional Captioner (DCC). The DCC model is further extended to an end-to-end system by simultaneously optimizing the visual recognition network, LSTM-based language model, and image captioning network with different sources in \cite{venugopalan2016captioning}.

Our model mainly focuses on the latter scenario, that incorporates the knowledge learnt from freely available unpaired object recognition data for novel object captioning. Different from previous methods which solely rely on the standard word-by-word sentence generation through a decoder RNN, we integrate the regular decoder RNN with copying mechanism which can simultaneously ``copy" the novel objects to the output sentence and the framework is trainable in an end-to-end fashion.

\begin{figure*}
\centering {\includegraphics[width=0.95\textwidth]{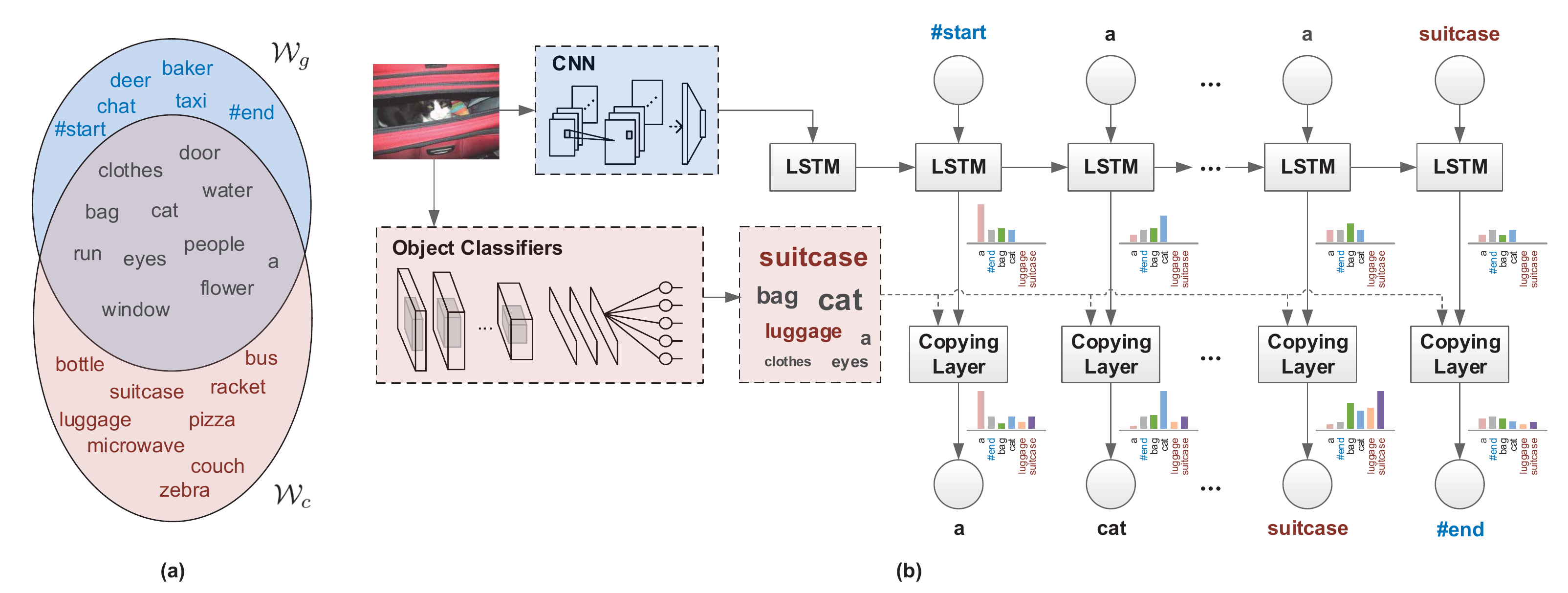}}
\caption{\small The overview of Long Short-Term Memory with Copying Mechanism (LSTM-C) for describing novel objects (better viewed in color). (a) $\mathcal {W}_g$ and $\mathcal {W}_c$ are the vocabularies on paired image-sentence dataset and unpaired object recognition dataset, respectively. (b) The image representation extracted by CNN is injected into LSTM at the initial time for standard word-by-word sentence generation. Meanwhile, the object classifiers learnt on unpaired object recognition dataset are utilized to detect the object candidates which are additionally incorporated into LSTM for directly ``copying" them into the output sentence, enabling the captioning for novel objects. To better leverage both generative mechanism for standard word-by-word sentence generation and our adopted copying mechanism, a copying layer is specially devised to integrate them in an end-to-end trainable architecture.}
\label{fig:figLC}
\vspace{-0.15in}
\end{figure*}

\section{Image Captioning with Copying Mechanism}\label{sec:IDC}
The main goal of our Long Short-Term Memory with Copying Mechanism (LSTM-C) framework is to describe novel objects in the output sentences by incorporating the copying mechanism into the decoding stage of image captioning. The overall training of LSTM-C is similar to regular CNN plus RNN systems by minimizing the energy loss which estimates the contextual relationships among the generated words in the decoding stage. Particularly, we measure the log probability of target word through not only the natural generation by generic RNN decoder, but also the direct ``copying" from the detected objects learnt on largely object recognition datasets, enabling the captioning for novel objects. The framework overview is shown in Figure~\ref{fig:figLC}.

In the following, we will first define the representations of images, the sequential words in sentence and the detected objects from images, followed by sequence modeling in image captioning. Next, to select words from novel objects and put them at proper places in the output sentence, we present the copying mechanism for image captioning from the viewpoint of rote memorization like a human being. Finally, the overall objective and optimization strategy of LSTM-C are presented in a CNN plus RNN framework. Technically, we devise a copying layer at the top of CNN plus RNN architecture, which incorporates both generative and copying mechanisms to optimize the whole network.

\subsection{Notation}
Suppose we have an image ${I}$ to be described by a textual sentence $\mathcal {S}$, where $\mathcal{S} = \{w_1, w_2, ..., w_{N_s}\}$ consisting of $N_s$ words. Let ${\bf{I}}\in {\mathbb{R}}^{D_v}$ and ${\bf{w}}_t\in {{\mathbb{R}}^{D_w}}$ denote the $D_v$-dimensional visual representations of the image ${I}$ and the $D_w$-dimensional textual features of the $t$-th word in sentence $\mathcal{S}$, respectively. As a sentence consists of a sequence of words, a sentence can be represented by a $D_w\times N_s$ matrix ${\bf{W}}\equiv [{\bf{w}}_1, {\bf{w}}_2,...,{\bf{w}}_{N_s}]$, with each word in the sentence as its column vector. The vocabulary for the paired image-sentence data is denoted as $\mathcal {W}_g$. Furthermore, we utilize the freely available object recognition datasets to train the object classifiers which will be injected into our CNN plus RNN system for novel object captioning. Let $\mathcal {W}_c$ denote the vocabulary for the unpaired object recognition dataset and the probability of image $I$ containing each object ${w_i}\in {\mathcal {W}_c}$ is represented as $\delta(w_i)$. More specifically, for the external images with single label (e.g., ImageNet \cite{ILSVRC15}), the standard CNN architecture \cite{Simonyan14} is adopted to train the object detectors, while for the image data with multiple objects (e.g., MSCOCO \cite{Lin:ECCV14}), we follow \cite{Fang:CVPR15} and learn the detectors by using the weakly-supervised approach of Multiple Instance Learning (MIL).

\subsection{Sequence Modeling in Image Captioning}
Inspired by the recent successes of probabilistic sequence methods leveraged in statistical machine translation \cite{Bahdanau14,Sutskever:NIPS14}, we aim to formulate our image captioning model in an end-to-end fashion based on RNN model which first encodes the given image into a fixed dimensional vector and then decodes it to the target output sentence consisting of sequential words. As such, given the image, the problem of sequence modeling for target sentence we exploit here can be generally formulated by minimizing the following energy loss function:
\begin{equation}\label{Eq:Eq1}\small
E({\bf {I}}, {\bf {W}}) = -\log {\Pr{({\bf {W}}|{\bf {I}})}},
\end{equation}
which is the negative log probability of the correct textual sentence given the visual image.

Since the model produces one word in the sentence at each time step, it is natural to apply chain rule to model the joint probability over the sequential words. Thus, the $\log$ probability of the sentence is given by the sum of the $\log$ probabilities over the word and can be expressed as
\begin{equation}\label{Eq:Eq2}\small
\log {\Pr{({\bf {W}}|{\bf {I}})}} =  \sum\limits_{t = 1}^{{N_s}} {\log \Pr\left( {\left. {{{\bf{w}}_t}} \right|{\bf{I}},{{\bf{w}}_0}, \ldots ,{{\bf{w}}_{t - 1}}} \right)}.
\end{equation}
By minimizing this loss, the contextual relationship among the words in the sentence can be guaranteed given the visual content of image.

We formulate this task as a variable-length sequence to sequence problem and model the parametric distribution $\Pr\left( {\left. {{{\bf{w}}_t}} \right|{\bf{I}},{{\bf{w}}_0}, \ldots ,{{\bf{w}}_{t - 1}}} \right)$ in Eq.(\ref{Eq:Eq2}) with LSTM, which is a widely used type of RNN in image/video captioning \cite{Vinyals14,Yao:arxiv16,Pan:CVPR16,Xu:CVPR16}. The vector formulas for a LSTM layer forward pass are given as below. For time step $t$, ${\bf{x}}^t$ and ${\bf{h}}^t$ are the input and output vector respectively, $\bf{T}$ are input weights matrices, $\bf{R}$ are recurrent weight matrices and $\bf{b}$ are bias vectors. Sigmoid $\sigma$ and hyperbolic tangent $\phi$ are element-wise non-linear activation functions. The dot product of two vectors is denoted with $\odot$. Given inputs ${\bf{x}}^t$, ${\bf{h}}^{t-1}$ and ${\bf{c}}^{t-1}$, the LSTM unit updates for time step $t$ are:
\begin{equation*}\label{Eq:Eqlstm3}\small
{{\bf{g}}^t} = \phi ({{\bf{T}}_g}{{\bf{x}}^t} + {{\bf{R}}_g}{{\bf{h}}^{t - 1}} + {{\bf{b}}_g}),~{{\bf{i}}^t} = \sigma ({{\bf{T}}_i}{{\bf{x}}^t} + {{\bf{R}}_i}{{\bf{h}}^{t - 1}} + {{\bf{b}}_i}),
\end{equation*}
\begin{equation*}\label{Eq:Eqlstm5}\small
~{{\bf{f}}^t} = \sigma ({{\bf{T}}_f}{{\bf{x}}^t} + {{\bf{R}}_f}{{\bf{h}}^{t - 1}} + {{\bf{b}}_f}),~~~{{\bf{c}}^t} = {\bf{g}}^t \odot {\bf{i}}^t + {\bf{c}}^{t-1} \odot {\bf{f}}^t,~~
\end{equation*}
\begin{equation*}\label{Eq:Eqlstm7}\small
{{\bf{o}}^t} = \sigma ({{\bf{T}}_o}{{\bf{x}}^t} + {{\bf{R}}_o}{{\bf{h}}^{t - 1}} + {{\bf{b}}_o}),~~~~{{\bf{h}}^t} = \phi ({\bf{c}}^t) \odot {\bf{o}}^t,~~~~~~~~
\end{equation*}
where ${{\bf{g}}^t}$, ${{\bf{i}}^t}$, ${{\bf{f}}^t}$, ${{\bf{c}}^t}$, ${{\bf{o}}^t}$, and ${{\bf{h}}^t}$ are cell input, input gate, forget gate, cell state, output gate, and cell output of the LSTM, respectively.

As mentioned above, the LSTM model is utilized to predict each word in the sentence given the image content and previous words. We inject the embedded image representation at the initial time to inform the whole memory cells in LSTM about the visual content. Given the image ${\bf{I}}$ and the corresponding sentence ${\bf{W}}\equiv [{\bf{w}}_0, {\bf{w}}_1,...,{\bf{w}}_{N_s}]$, the LSTM updating procedure is as following:
\begin{equation}\label{Eq:Eq9}\small
{{\bf{x}}^{-1}} = {{\bf{T}}_{I}}{\bf{I}},~~~~~~
\end{equation}
\begin{equation}\label{Eq:Eq10}\small
{{\bf{x}}^t} = {{\bf{T}}_s}{{\bf{w}}_t},~~~t \in \left\{ {0, \ldots ,{N_s}-1} \right\}~,
\end{equation}
\begin{equation}\label{Eq:Eq11}\small
{{\bf{h}}^{t}} = f\left( {{{\bf{x}}^t}} \right),~~t \in \left\{ {0, \ldots ,{N_s}-1} \right\},
\end{equation}
where $D_e$ is the dimensionality of LSTM input, and ${{\bf{T}}_I} \in {{\mathbb{R}}^{{D_e} \times {D_v}}}$ and ${{\bf{T}}_s} \in {{\mathbb{R}}^{{D_e} \times {D_w}}}$ are the transformation matrices for image representation and textual feature of word, respectively, and $f$ is the updating function within LSTM unit. Please note that for the input sentence ${\bf{W}} \equiv [ {{{\bf{w}}_0}, \ldots,{{\bf{w}}_{{N_s}}}} ]$, we take ${{\bf{w}}_0}$ as the start sign word to inform the beginning of sentence and ${{\bf{w}}_{{N_s}}}$ as the end sign word which indicates the end of sentence, both of the special sign words are included in the existing vocabulary $\mathcal {W}_g$ for the paired image-sentence data. More specifically, at the initial encoding step, the image representation is transformed as the input for LSTM, and then in the next decoding steps, word embedding ${{\bf{x}}^t}$ will be input into the LSTM along with the previous step's hidden state ${{\bf{h}}^{t-1}}$.

In the decoding stage, given the LSTM cell output ${{\bf{h}}^{t}}$ at the $t$-th time step, the widely adopted method for next word prediction is the generative mechanism \cite{Bahdanau14} which calculates the corresponding probability of generating any target word $w_{t+1}$ as
\begin{equation}\label{Eq:Eq12}
{\Pr}_t^g\left( {{w_{t + 1}}} \right) = {{\bf{w}}_{t+1}^\top}{{\bf{M}}_g}{{\bf{h}}^t},
\end{equation}
where $D_h$ is the dimensionality of LSTM output and ${{\bf{M}}_g} \in {{\mathbb{R}}^{{D_w} \times {D_h}}}$ is the transformation matrix for textual features of word in the generative mechanism. For the standard word-by-word sentence generation model, a softmax function is applied after the probabilities measured by generative mechanism to produce a normalized probability distribution over all the words in the vocabulary $\mathcal {W}_g$.

\subsection{Copying Mechanism}
The copying mechanism has been shown effective for sequence learning \cite{gu2016incorporating} to address the out-of-vocabulary (OOV) problem in text summarization. The mechanism is regarded as the rote memorization in language processing of human being that directly ``copying" existing segments in the input sequence to target sequence. Similar in spirit, we extend the copying mechanism in image captioning to directly ``copying" the appropriate objects from the detected candidates in image to compose the output sentence, especially for novel objects which never appear in paired image-sentence data, enabling the novel object captioning. Specifically, at the $t$-th decoding step, we directly take the similarity between any word $w_{t+1}$ in $\mathcal {W}_c$ and the corresponding LSTM cell output ${{\bf{h}}^{t}}$ as the probability for ``copying" the target word $w_{t+1}$ to the target sentence, which is calculated as
\begin{equation}\label{Eq:Eq13}
{\Pr}_t^c\left( {{w_{t + 1}}} \right) =\varphi \left( {{\bf{w}}_{t + 1}^\top{{\bf{M}}_c}} \right){{\bf{h}}^t}{\delta \left( {{w_{t + 1}}} \right)} ,
\end{equation}
where ${{\bf{M}}_c} \in {{\mathbb{R}}^{{D_w} \times {D_h}}}$ is the transformation matrix for mapping textual features of word in the copying mechanism and $\varphi$ is element-wise non-linear activation function. It is also worth noticing that we additionally incorporate the object classification score $\delta \left( {{w_{t + 1}}}\right)$ into the formulation of ``copying" probability since this classification score reflects the chance of the object appeared in the image. The underlying assumption is that in addition to the effect of LSTM cell output, the larger the classification score of this word in the image, the higher the probability for ``copying" this word in the target sentence.

\subsection{LSTM with Copying Mechanism}
Unlike the existing image captioning approaches which always model the sequence learning with generative mechanism for sentence generation, our proposed LSTM-C architecture further incorporates the copying mechanism into LSTM at the decoding stage to describe novel objects in sentence. In particular, given the output of LSTM cell at each decoding step, we utilize both generative and copying mechanisms simultaneously to measure the probability of generating any target word. As the vocabulary $\mathcal {W}_c$ of copying mechanism is derived from external image data, it may include the words which are not present in the vocabulary $\mathcal {W}_g$ of image-sentence data, making the copying mechanism able to copy such novel objects to the output sentence. In this case, we directly consider the probability of copying mechanism in Eq.(\ref{Eq:Eq13}) as the final probability of generating these novel objects. Similarly, for the words that only belong to $\mathcal {W}_g$, the final probabilities of them fully depend on their corresponding probabilities of generative mechanism in Eq.(\ref{Eq:Eq12}). In terms of the overlapping words between $\mathcal {W}_g$ and $\mathcal {W}_c$, we linearly fuse the probabilities from both generative and copying mechanisms as the final output probabilities. Hence, at the $t$-th decoding step, the final output probability ${{\Pr}_t}\left( {{w_{t + 1}}} \right)$ of any target word $w_{t+1}$ is defined as follows:
\begin{equation}\label{Eq:Eq14}\scriptsize
{{\Pr}_t}\left( {{w_{t + 1}}} \right) = \left\{ {\begin{array}{*{20}{c}}
~~~~~~~~~~~~~~~~~{\frac{1}{K}{e^{\Pr _t^g\left( {{w_{t + 1}}} \right)}},~~~~~~~~~~~~~~~~{w_{t + 1}} \in {\mathcal {W}_g} \cap \overline {{\mathcal {W}_c}} }\\\\
{\frac{\lambda }{K}{e^{\Pr _t^g\left( {{w_{t + 1}}} \right)}} + \frac{{1 - \lambda }}{K}{e^{\Pr _t^c\left( {{w_{t + 1}}} \right)}},{w_{t + 1}} \in {\mathcal {W}_g} \cap {\mathcal {W}_c}}\\\\
~~~~~~~~~~~~~~~~~~{\frac{1}{K}{e^{\Pr _t^c\left( {{w_{t + 1}}} \right)}},~~~~~~~~~~~~~~~~{w_{t + 1}} \in \overline {{\mathcal {W}_g}}  \cap {\mathcal {W}_c}}\\\\
{~~~~~~~~~~~~~~~~~~~0,~~~~~~~~~~~~~~~~~~~~~~~~~~~otherwise}
\end{array}} \right.,
\end{equation}
where $\lambda$ is the tradeoff parameter between the two mechanisms and $K$ is the softmax normalization term.

Accordingly, we define our energy loss function in training stage for each image-sentence pair as follows:
\begin{equation}\label{Eq:Eq15}\small
E(I, \mathcal{S}) = - \sum\limits_{t = 0}^{{N_s-1}} {\log {\Pr}_t({\bf{w}}_{t+1} )}.
\end{equation}

Let $N$ denote the number of image-sentence pairs in the training set, we have the following optimization problem:
\begin{equation}\label{Eq:Eq16}\small
\begin{array}{l}
\mathop {\min }\limits_{{{\bf{T}}_I},{{\bf{T}}_s},{{\bf{M}}_g},{{\bf{M}}_c},\theta } \frac{1}{N}\sum\limits_{i = 1}^N { E(I^{(i)}, \mathcal{S}^{(i)})} \\
~~~~~~~~~~~~~+ \left\| {{{\bf{T}}_I}} \right\|_2^2 + \left\| {{{\bf{T}}_s}} \right\|_2^2 + \left\| {{{\bf{M}}_g}} \right\|_2^2 + \left\| {{{\bf{M}}_c}} \right\|_2^2 + \left\| \theta  \right\|_2^2
\end{array},
\end{equation}
where the first term is the overall energy loss, and the rest are regularization terms for image embedding, textual embedding for LSTM input, textual embedding in generative mechanism, textual embedding in copying mechanism, and LSTM, respectively. Moreover, following \cite{venugopalan2016captioning}, we also implicitly integrate the overall energy loss with text-specific loss on external sentence data for maintaining the model's ability to address novel objects among sentences.

To solve the optimization according to overall loss objective in Eq.(\ref{Eq:Eq16}), we design a copying layer at the top of LSTM with two textual embedding parameters for generative and copying mechanisms. During training, this copying layer measures the output probability for each word considering both generative and copying mechanisms as defined in Eq.(\ref{Eq:Eq14}), followed by a softmax normalization operation for overall optimization.

In the testing stage for sentence generation, we choose the word among the combination vocabulary of ${\mathcal {W}_g}$ and ${\mathcal {W}_c}$ with maximum probability at each time step and set its embedded textual feature as LSTM input for the next time step until the end sign word is outputted.

\section{Experiments}\label{sec:EX}
We evaluate and compare our proposed LSTM-C with state-of-the-art approaches by conducting novel object captioning task on two image datasets, i.e., the held-out Microsoft COCO Caption dataset (held-out MSCOCO) \cite{Hendricks:CVPR16} which is a subset of MSCOCO dataset \cite{Lin:ECCV14} and ImageNet \cite{ILSVRC15}, a large-scale object recognition dataset.

\subsection{Datasets}
\paragraph{Held-out MSCOCO.} The held-out MSCOCO consists of a subset of MSCOCO which excludes all the image-sentence pairs that contain at least one of eight specific objects in MSCOCO. It is worth noting that following \cite{Hendricks:CVPR16}, the eight specific objects are chosen through the clustering over all the 80 objects in MSCOCO segmentation challenge and each cluster excludes one object, resulting in the final eight novel objects for evaluation: ``bottle," ``bus," ``couch," ``microwave," ``pizza," ``racket," ``suitcase," and ``zebra." For this subset, there are five human-annotated descriptions per image. As the annotations of the official testing set are not publicly available and thus following \cite{Hendricks:CVPR16}, we split the MSCOCO validation set into two: 50\% for validation and the other 50\% for testing. For the experiments on held-out MSCOCO, the object classifiers for copying mechanism are trained with all the MSCOCO training images including the eight novel objects and the LSTM for sequence modeling is pre-trained with all the sentences in MSCOCO training set, while the entire CNN plus RNN system are optimized with the paired image-sentence data only from held-out MSCOCO training set. The testing set of held-out MSCOCO is then utilized to evaluate the ability of our LSTM-C model to describe the eight novel objects.

\paragraph{ImageNet.} We also conduct our experiments on the large-scale object recognition dataset, i.e., ImageNet, for evaluation. Similar to \cite{venugopalan2016captioning}, a subset from ImageNet with 634 different objects which are not present in the MSCOCO dataset is adopted in our experiments. In particular, about 75\% of images in each class are exploited for training and the rest are utilized for testing, resulting in the training and testing set with 493,519 and 164,820 images, respectively. For the experiments on ImageNet, we train the object classifiers for copying mechanism purely on the ImageNet training set and pre-train the LSTM part with all the sentences in MSCOCO training set. In terms of the entire CNN plus RNN system, it is optimized with the paired image-sentence data in MSCOCO training set. Since none of the objects in this subset of ImageNet is addressed in the paired image-sentence data, we generate sentences for images in the testing set of ImageNet and empirically evaluate the ability of our LSTM-C model to describe the 634 novel objects.

\subsection{Experimental Settings}
\paragraph{Features and Parameter Settings.}
For image representations, we take the output of 4,096-way fc7 layer from 16-layer VGG \cite{Simonyan14} pre-trained on Imagenet ILSVRC12 dataset \cite{ILSVRC15}. Each word in the sentence is represented as the combined vector of embedded one-hot representation and Glove \cite{pennington2014glove} representation. For the paired image-sentence data (e.g., MSCOCO), we select the 1,000 most common words on MSCOCO as the objects and train the corresponding object classifiers with MIL model \cite{Fang:CVPR15} purely on the training data of MSCOCO. The MIL model is mainly designed based on a Fully Convolutional Network (FCN) extended from 16-layer VGG. For the unpaired object recognition data (e.g., ImageNet), 634 object classifiers are trained by directly fine-tuning the 16-layer VGG pre-trained on Imagenet ILSVRC12 dataset. The dimensionality of the input and hidden layers in LSTM are both set to 1,024. The tradeoff parameter $\lambda$ leveraging both generative and copying mechanisms is empirically set to 0.2. The sensitivity of $\lambda$ will be discussed later.

\paragraph{Implementation Details.}
We mainly implement our image captioning models based on Caffe \cite{Jia:MM14}, which is one of widely adopted deep learning frameworks. In particular, the initial learning rate and mini-batch size is set as 0.01 and 1,024, respectively. The entire CNN plus RNN system in our LSTM-C is trained for 50 epoches on both datasets or we stop the training until the performance has no longer improvement on the corresponding validation set.

\paragraph{Evaluation Metrics.}
For quantitative evaluation of our proposed model on held-out MSCOCO, we adopt the most common caption metric, i.e., METEOR \cite{Banerjee:ACL05}, to evaluate description quality which computes unigram precision and recall against all ground truth sentences with some pre-processing on WordNet synonyms and stemmed tokens. However, as pointed in \cite{Hendricks:CVPR16}, it is still possible to achieve high METEOR scores without mentioning the novel objects. Hence, to fully validate the model's ability of describing novel objects, F1-score is exploited as another evaluation metric, which determines whether the specific novel object is mentioned in the generated descriptions for the images containing that novel object. All the metrics above are computed by using the codes\footnote{\url {https://github.com/LisaAnne/DCC}} released by \cite{Hendricks:CVPR16} for fair comparison. To evaluate our model on ImageNet without any ground truth sentences, we utilize another two metrics for novel object captioning task: describing novel objects (Novel) \cite{venugopalan2016captioning} and Accuracy \cite{venugopalan2016captioning} scores. The Novel score measures the percentage of all the 634 novel objects mentioned in generated descriptions, i.e., for each novel object, the model should incorporate it into at least one sentence for the ImageNet image with this object. For the Accuracy score of each novel object, it represents the percentage of images belonging to this novel object which can be described correctly by addressing that novel object in the sentences. The Accuracy score is finally averaged over all the 634 novel objects.

\begin{table*}[!tb]\small
\centering
\caption{Per-object F1, averaged F1 and METEOR scores of our proposed model and other state-of-the-art methods on held-out MSCOCO dataset for novel object captioning. All values are reported as percentage (\%).}
\label{table:FMCOCO}
\begin{tabular}{|l|cccccccc|c|c|}\hline
~~Model&~F1$_\text{bottle}$~&~F1$_\text{bus}$~&~F1$_\text{couch}$~&~F1$_\text{microwave}$~&~F1$_\text{pizza}$~&
~F1$_\text{racket}$~&~F1$_\text{suitcase}$~&~F1$_\text{zebra}$~&~F1$_\text{average}$~&~METEOR~
\\ \hline\hline
~~LRCN \cite{Donahue14} &0 &0 &0 &0 &0 &0 &0 &0 &0 &19.33\\
~~DCC \cite{Hendricks:CVPR16} & 4.63 & 29.79& \textbf{45.87} & 28.09 & 64.59& 52.24& 13.16& 79.88& 39.78 &21 \\
~~NOC \cite{venugopalan2016captioning} & & & &&&&&&&\\
~~~-(One hot)       &16.52 &68.63 &42.57 &32.16 &67.07 &61.22 &31.18 &88.39 &50.97 &20.7\\
~~~-(One hot+Glove) &14.93 &68.96 &43.82 &\textbf{37.89} &66.53 &65.87 &28.13 &88.66 &51.85 &20.7\\
~~\textbf{LSTM-C}& & & &&&&&&&\\
~~~-(One hot)       &29.07 &64.38 &26.01 &26.04 &\textbf{75.57} &66.54 &\textbf{55.54} &\textbf{92.03}     &54.40 &22\\
~~~-(One hot+Glove) &\textbf{29.68} &\textbf{74.42}  &38.77  &27.81  &68.17  &\textbf{70.27}                &44.76  &91.4                &\textbf{55.66}& \textbf{23}\\ \hline
\end{tabular}
\vspace{-0.1in}
\end{table*}

\subsection{Compared Approaches}
To empirically verify the merit of our LSTM-C model, we compared the following state-of-the-art methods, including both regular image captioning and novel object captioning approaches.
\begin{itemize}
  \item Long-term Recurrent Convolutional Networks (LRCN) \cite{Donahue14}: LRCN is one of the basic RNN-based image captioning models which inputs both visual image and previous word into LSTM at each time step for sentence generation. As a regular image captioning model without any mechanism for considering novel objects, LRCN is trained only on the paired image-sentence data without any novel objects.
  \item Deep Compositional Captioner (DCC) \cite{Hendricks:CVPR16}: DCC firstly pre-trains lexical classifier and language model with external unpaired data, and then integrates both two parts to learn an improved caption model trained with paired image-sentence data. Finally, DCC explicitly transfers the knowledge of semantically related objects to compose the descriptions with novel objects.
  \item Novel Object Captioner (NOC) \cite{venugopalan2016captioning}: Proposed most recently, NOC extends DCC by jointly optimizing the three parts: visual recognition network, LSTM-based language model, and image captioning network in an end-to-end manner. Please note that for fair comparison with LRCN and DCC which utilize one hot vector as their word representations, we include two runs, i.e., NOC (One hot) and NOC (One hot+Glove) which are our implementations of NOC. The word representations in the latter one are the combination of the embedded one hot vector and Glove vector.
  \item Long Short-Term Memory with Copying Mechanism (LSTM-C): We design two runs, i.e., LSTM-C (One-hot) and LSTM-C (One hot+Glove), for our proposed end-to-end architecture for novel object captioning.
\end{itemize}

\subsection{Performance Comparison}
We first conduct the experiment on held-out MSCOCO to examine how our LSTM-C model work on describing the eight novel objects. Then, to further verify the scalability of our proposed model, the second experiment is performed on ImageNet to describe hundreds of novel objects that outside of the paired image-sentence data.

\textbf{Evaluation on held-out MSCOCO.} Table \ref{table:FMCOCO} shows the performances of compared six models on held-out MSCOCO dataset. Overall, the results across two general evaluation metrics (averaged F1 and METEOR scores) consistently indicate that our proposed LSTM-C exhibits better performance than all the state-of-the-art techniques including regular image captioning model (LRCN) and two novel object captioning systems (DCC and NOC). In particular, by additionally utilizing external unpaired data for training, all the latter five novel object captioning models outperform the regular image captioning model LRCN on both description quality and novelty. There is a significant performance gap between DCC and LSTM-C (One hot). Although both runs involve the utilization of external image data, they are fundamentally different in the way that DCC leverages explicit transfer mechanism for recognizing novel objects and cannot be trained end-to-end, and LSTM-C (One hot) implicitly addresses the novel objects for sentence generation with copying mechanism in an end-to-end manner. Moreover, by incorporating copying mechanism to standard word-by-word sentence generation model, LSTM-C (One hot) leads to a performance boost against NOC (One hot), indicating that the generative mechanism and copying mechanism are complementary and thus have mutual reinforcement for novel object captioning. Another observation is that when combining the word representations from embedded one hot vector and Glove vector, LSTM-C (One hot+Glove) further increases the performance.

\begin{table}[!tb]
\centering
\caption{Novel, F1 and Accuracy scores of our proposed model and other state-of-the-art methods on ImageNet dataset. All values are reported as percentage (\%).}
\label{table:NFIM}\small
\begin{tabular}{|l|c|c|c|}\hline
Model~~&~~Novel~~&~~F1~~&~~Accuracy~~
\\ \hline\hline
NOC (One hot+Glove) \cite{venugopalan2016captioning} & & & \\
~~~~-MSCOCO  &69.08 & 15.63 & 10.04\\
~~~~-BNC\&Wiki &87.69  & 31.23 &21.96\\
LSTM-C (One hot+Glove) & & & \\
~~~~-MSCOCO  &72.08 & 16.39 & 11.83\\
~~~~-BNC\&Wiki &89.11  & 33.64 &31.11\\\hline
\end{tabular}
\vspace{-0.1in}
\end{table}

Table \ref{table:FMCOCO} also details the F1 scores for all the eight novel objects. Among all the novel objects, our proposed LSTM-C achieves the best performance for describing six novel objects, followed by DCC and NOC for one object, respectively. The improvements can be generally expected by additionally incorporating copying mechanism in sequence learning except ``couch" and ``microwave" objects. This is not surprise because such novel objects always have high visual similarity with other objects (e.g., ``bed" for ``couch" and ``oven" for ``microwave") and thus are not easy to be detected precisely, making LSTM-C fail to copy them to the output sentences.

\textbf{Evaluation on ImageNet.} Table \ref{table:NFIM} summarizes the experimental results on ImageNet dataset. By only adopting the MSCOCO as the training data for the CNN plus RNN system, our LSTM-C (One hot+Glove) makes the relative improvement over NOC (One hot+Glove) by 4.3\%, 4.9\% and 17.8\% in Novel, F1 and Accuracy, respectively. The results basically indicate the advantage of exploiting both generative and copying mechanisms in the CNN plus RNN system for novel object captioning, even when scaling into ImageNet images with hundreds of novel objects. Moreover, following \cite{venugopalan2016captioning}, we also include the external unpaired text data (i.e., British National Corpus and Wikipedia) in our LSTM-C (One hot+Glove) and performance improvements are further observed.

\begin{figure}[!tb]
\centering {\includegraphics[width=0.45\textwidth]{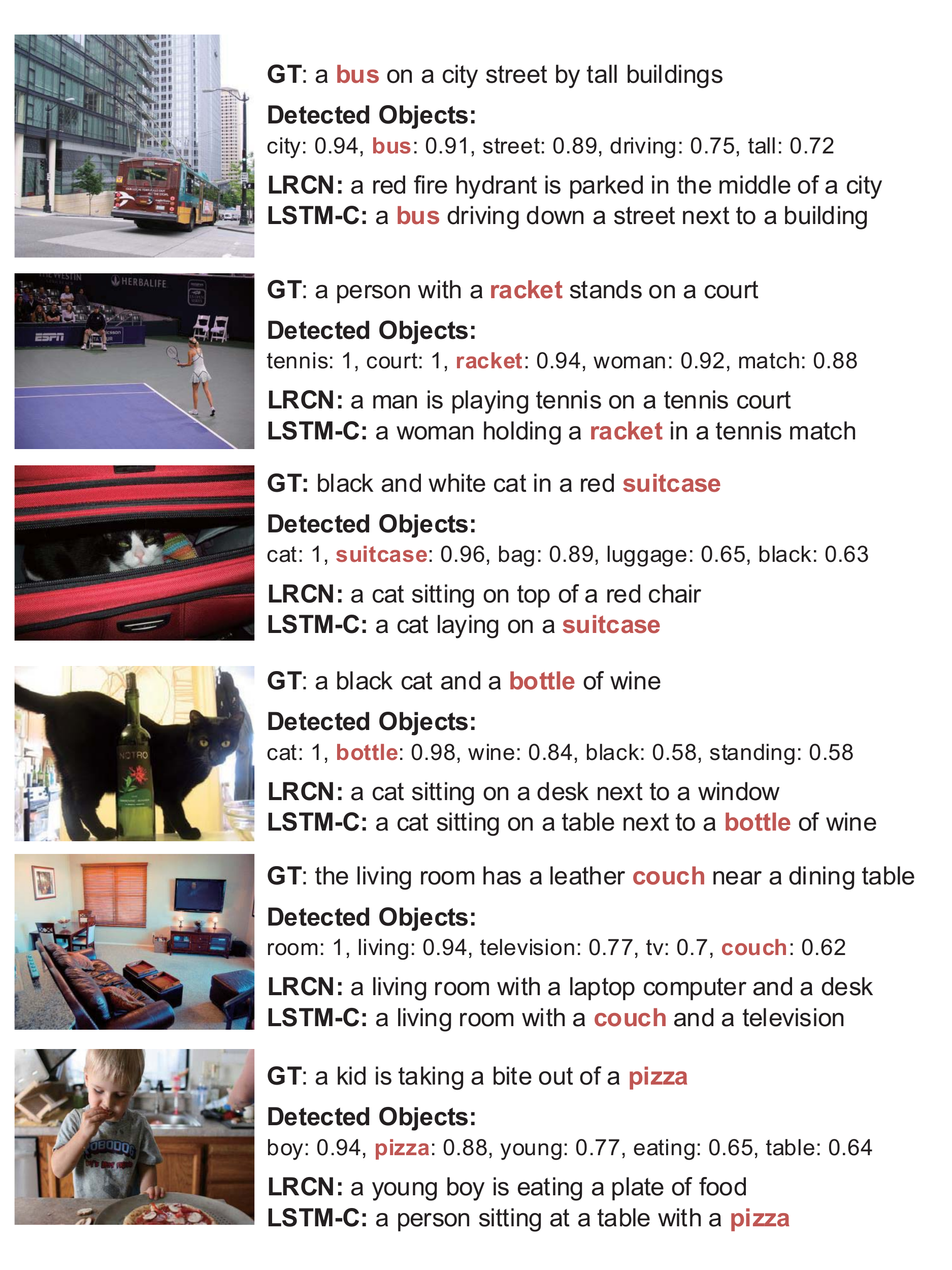}}
\caption{\small Objects and sentence generation results on held-out MSCOCO. The detected objects are predicted by MIL model in \cite{Fang:CVPR15}, and the output sentences are generated by 1) Ground Truth (GT): one ground truth sentence, 2) LRCN and 3) our LSTM-C.}
\label{fig:figRCOCO}
\vspace{-0.1in}
\end{figure}

\begin{figure}
\centering {\includegraphics[width=0.42\textwidth]{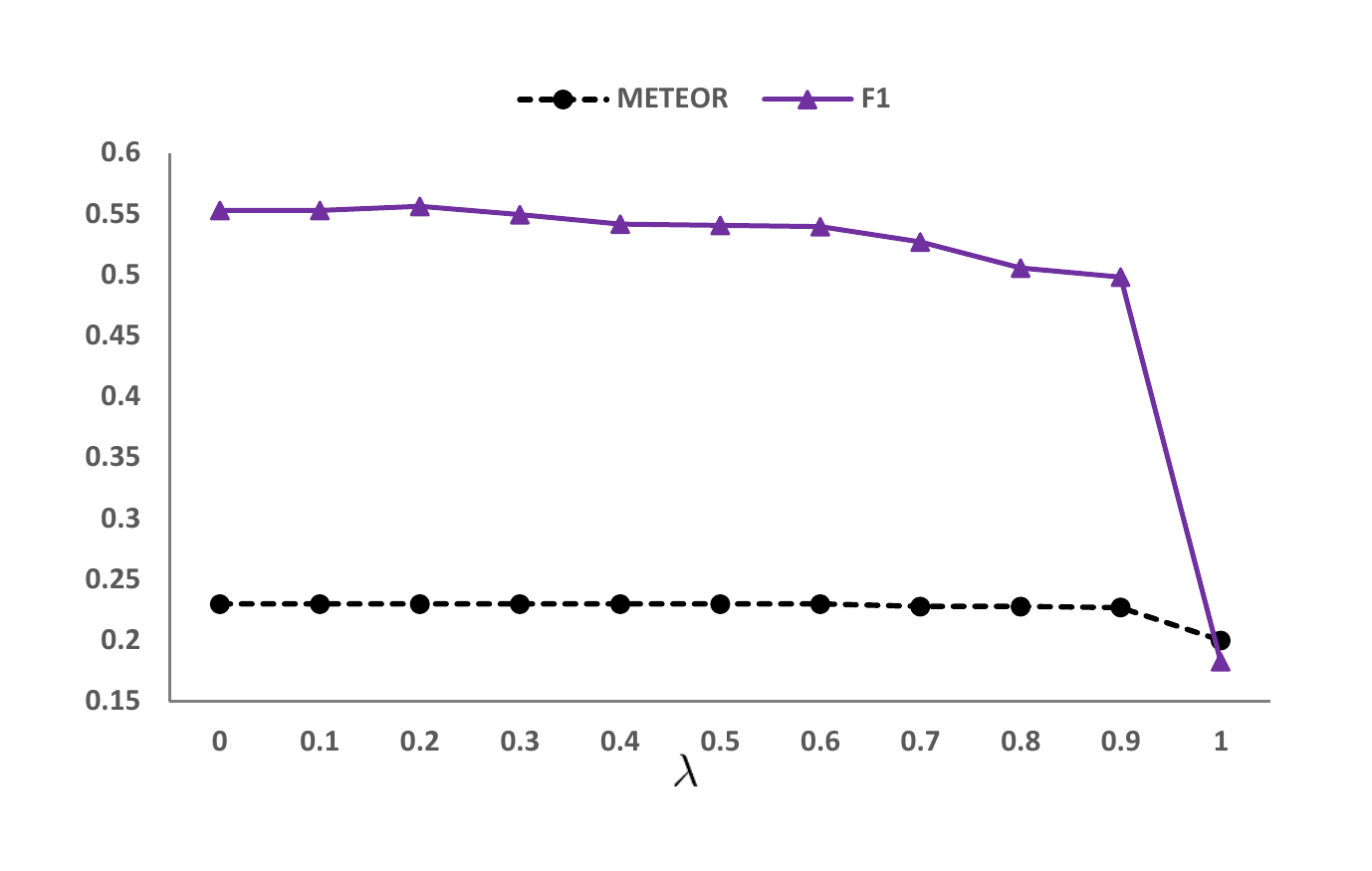}}
\caption{The effect of the tradeoff parameter $\lambda$ in our LSTM-C (One hot+Glove) framework on held-out MSCOCO.}
\label{fig:figLambda}
\vspace{-0.12in}
\end{figure}

\begin{figure}[!tb]
\centering {\includegraphics[width=0.45\textwidth]{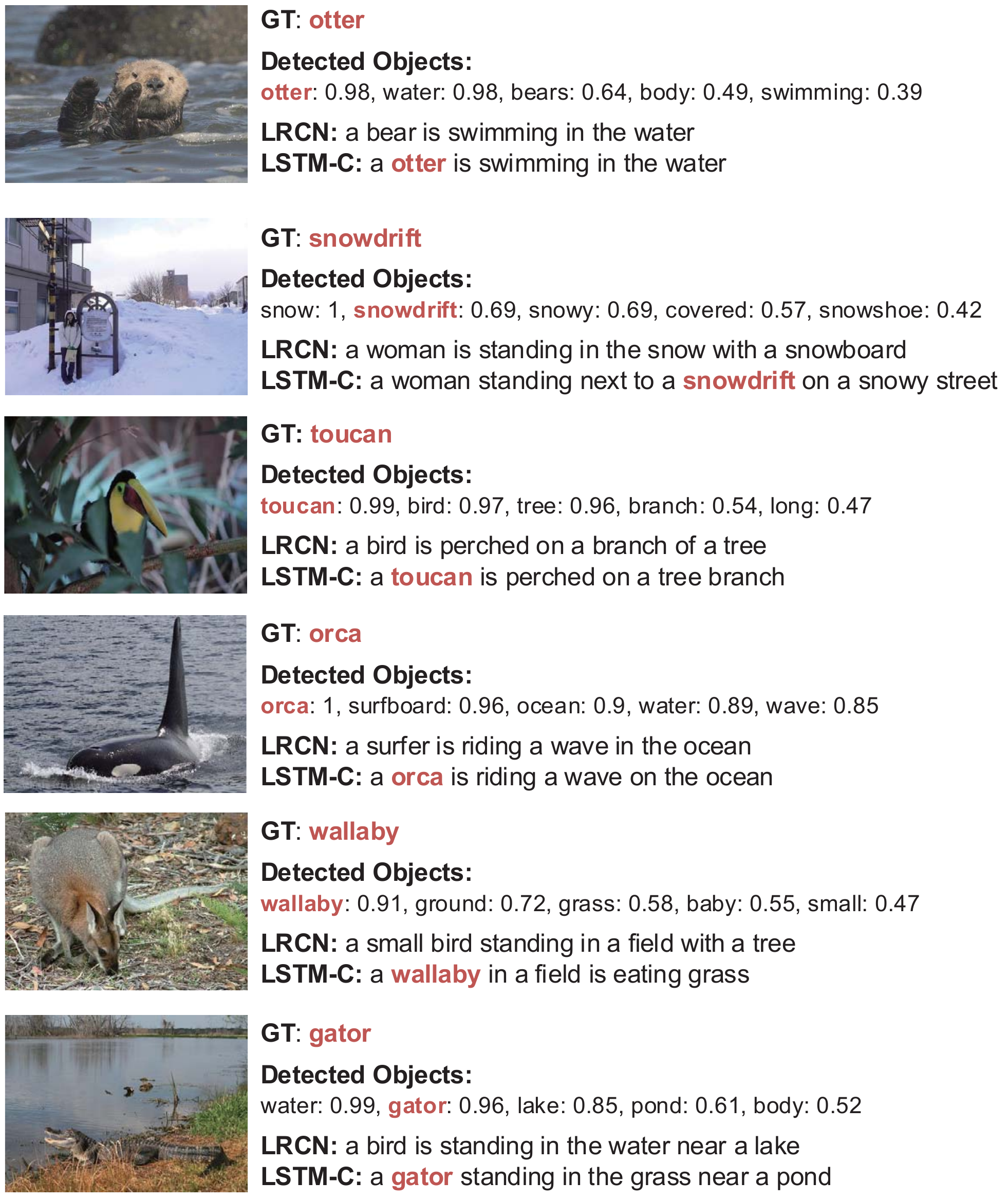}}
\caption{Objects and sentence generation results on ImageNet. GT denotes the ground truth object. The detected objects are predicted by the standard CNN architecture \cite{Simonyan14}, and the output sentences are generated by 1) LRCN and 2) our LSTM-C.}
\label{fig:figImageNet}
\vspace{-0.15in}
\end{figure}

\textbf{Qualitative analysis.} Figure \ref{fig:figRCOCO} and Figure \ref{fig:figImageNet} shows a few sentence examples generated by different methods, the detected objects and human-annotated ground truth on held-out MSCOCO and ImageNet dataset, respectively. From these exemplar results, it is easy to see that all of these captioning models can generate somewhat relevant sentences on both datasets, while our proposed LSTM-C can predict the novel objects by incorporating copying mechanism for image captioning. For example, compared to object term ``hydrant" in the sentence generated by LRCN, ``bus" in our LSTM-C is more precise to describe the image content in the first image on held-out MSCOCO dataset, since the novel object ``bus" is among the top object candidates and directly copied to the output sentence at the decoding stage.

\subsection{Analysis of the Tradeoff Parameter $\lambda$}
To clarify the effect of the tradeoff parameter $\lambda$ in Eq.(\ref{Eq:Eq14}), we illustrate the performance curves with different tradeoff parameters in Figure \ref{fig:figLambda}. As shown in the figure, we can see that the performance curves of F1 and METEOR scores are both relatively smooth when $\lambda$ varies in a range from 0 to 0.6. Specifically, the best performance is achieved when $\lambda$ is about 0.2. Furthermore, when the $\lambda$ increases more than 0.6, the F1 score begins to drop significantly, again demonstrating the importance of copying mechanism in our LSTM-C for describing novel objects.

\section{Discussions and Conclusions}\label{sec:CON}
We have presented Long Short-Term Memory with Copying Mechanism (LSTM-C) framework which leverages external visual recognition for image captioning. Particulary, we study the problem of predicting novel objects in image caption by integrating the detected objects with copying mechanism. To verify our claim, we have devised an end-to-end architecture to accommodate the standard word-by-word sentence generation by LSTM and the mechanism of copying from detected objects. Experiments conducted on MSCOCO image captioning and ImageNet datasets validate our proposal and analysis. Performance improvements are clearly observed when comparing to other novel object captioning techniques.

Our future works are as follows. First, more objects will be learnt on large-scale image benchmarks, e.g., YFCC-100M dataset, and integrated into our LSTM-C architecture. We will further analyze the impact of different sources involved. Second, how to apply our proposal to video domain is also worth trying.

{\small
\bibliographystyle{ieee}
\bibliography{egbib}
}

\end{document}